\documentclass{Interspeech2024}

\usepackage{subfig}
\usepackage{multirow}
\usepackage{booktabs}

\interspeechcameraready

\title{Out-of-distribution generalisation in spoken language understanding}

\name{Dejan}{Porjazovski}
\name{Anssi}{Moisio}
\name{Mikko}{Kurimo}

\address{Aalto University, Finland}
\email{dejan.porjazovski@aalto.fi, anssi.moisio@aalto.fi, mikko.kurimo@aalto.fi}

\keywords{spoken language understanding, generalisation, out-of-distribution, model interpretability}

\begin{document}
% Make the table font 8pt
\AtBeginEnvironment{tabular}{\fontsize{8pt}{10pt}\selectfont}

\maketitle
 
\begin{abstract}
Test data is said to be out-of-distribution (OOD) when it unexpectedly differs from the training data, a common challenge in real-world use cases of machine learning. Although OOD generalisation has gained interest in recent years, few works have focused on OOD generalisation in spoken language understanding (SLU) tasks. To facilitate research on this topic, we introduce a modified version of the popular SLU dataset SLURP, featuring data splits for testing OOD generalisation in the SLU task. We call our modified dataset SLURP For OOD generalisation, or SLURPFOOD. Utilising our OOD data splits, we find end-to-end SLU models to have limited capacity for generalisation. Furthermore, by employing model interpretability techniques, we shed light on the factors contributing to the generalisation difficulties of the models. To improve the generalisation, we experiment with two techniques, which improve the results on some, but not all the splits, emphasising the need for new techniques.
\end{abstract}

\section{Introduction}
Spoken language understanding (SLU) systems are the backbone of human-computer interaction devices that need to understand the meaning of the utterance before taking an action. When these models are deployed for real-world use, their performance should be consistent even when presented with out-of-distribution (OOD) data that differs from the training distribution in an unpredictable manner \cite{liu2021towards}.

Multiple types of OOD generalisation capacities are desirable but not necessarily achieved by neural SLU systems. \emph{Length generalisation} (see e.g. \cite{pmlr-v202-ray-chowdhury23b,zhou2023algorithms}) is the capacity to process sequences that are longer (or shorter) than those seen in the training set. \emph{Out-of-vocabulary} (OOV) generalisation is needed when the test set includes words or other units not seen in the training set \cite{lugosch19_interspeech, palogiannidi2020end}. A third type is \emph{compositional generalisation} (CG) \cite{fodor1988connectionism, hupkes2020compositionality}, required when test samples combine familiar units in novel ways. For example, novel combinations of slot types in a slot filling task has been shown to pose difficulties for SLU systems \cite{broscheit2022distributionally, ray23_interspeech}. More generally, neural (NLP) systems have been found often to fail in tasks that require CG \cite{lake2018generalization, keysers2019measuring, yao2022structural}, although they do have some capacity for CG \cite{lepori2023break}. In addition to the generalisation types applicable to both text and audio sequences, some types are specific to audio-based models, such as diverse accents \cite{viglino19_interspeech}, different age groups \cite{potamianos2003robust}, and various acoustic environments \cite{haeb2020far}.

The traditional pipeline SLU systems consist of two sub-modules: an automatic speech recognition (ASR) system that generates transcripts and a separate text-based natural language understanding (NLU) system that extracts meaning from the generated transcripts \cite{gupta2005t, ray23_interspeech}. Previous studies that explore generalisation in SLU employ a pipeline system, focusing on the transcripts instead of the original audio \cite{gaspers2022temporal, ray23_interspeech}. As these studies focus on the text-based language understanding task, the proposed evaluation sets are mostly text-only data, and as such, they can not always be applied to the end-to-end (E2E) SLU models that rely solely on audio \cite{serdyuk2018towards, haghani2018audio, palogiannidi2020end}. The main reason is that text is easy to segment due to word boundaries, whereas adding or removing segments of an audio signal corresponding to certain tokens in the transcript is challenging. Moreover, the audio models' generalisation criteria can differ from those of text-based models.

Despite these challenges, a few previous works have evaluated the generalisation capabilities of E2E SLU models. In \cite{lugosch19_interspeech}, a system was trained on three specific phrases and tested with also a new phrase, assessing the OOV generalisation. Similarly, in \cite{palogiannidi2020end}, E2E SLU models were tested on unique wordings not seen in training. The findings revealed a notable increase in error rates as the number of unseen wordings increased. These studies, however, predominantly focus on a singular type of generalisation. The authors of \cite{arora21_interspeech} noted that generalisation in SLU tasks may be divided into two types: generalising to diverse speakers and generalising to diverse phrases (e.g., n-grams). These two types may be orthogonal and thus can be assessed separately. We argue, in the same vein, that there are multiple types of generalisation that SLU models should ideally achieve. However, our data splitting methods differ from those in \cite{arora21_interspeech} in multiple ways. The n-gram-based splits in \cite{arora21_interspeech} may result in OOV words in the test set, or possibly novel combinations (n-grams) of familiar words, requiring compositional generalisation. We distinguish OOV splits from CG splits, allowing for a more fine-grained evaluation. Moreover, we focus on slightly more abstract units than n-grams, namely the scenario and action labels. Our methods and resources therefore aim to complement the previous works, adding to the collection of resources enabling development of more robust SLU systems.

We introduce OOV, CG, and microphone mismatch splits of the SLURP dataset \cite{bastianelli-etal-2020-slurp}, which we call SLURP For OOD generalisation (SLURPFOOD). Utilising these splits, we assess SLU models' capacity for OOD generalisation. As a benchmark, we report results for pre-trained self-supervised models. We also leverage a model interpretability method
(Integrated Gradients \cite{sundararajan2017axiomatic}) to determine the underlying causes contributing to the limited generalisation of the systems. To improve the generalisation, we explore two techniques: TOPK \cite{kawaguchi2020ordered} and segmented processing. We make the SLURPFOOD splits, the models, and the code for reproducing all the experiments publicly available\footnote{https://github.com/aalto-speech/slurpfood}.

%%% Data stats %%%
\begin{table*}[htb]
    \footnotesize
    \caption{Statistics for the proposed split configurations. OOV is out-of-vocabulary, CG is compositional generalisation, and  DA is Double-Action. The mic mismatch split is not included due to its similarity to the original splits. The table shows class overlaps (intersections of class sets) between the train, dev, and test subsets for each label type (scenario, action, intent). In the brackets are the similarities $C_\alpha$ (see Section~\ref{sec:dbca}) of the frequency distributions of labels in the different subsets.}
    \label{tab:split_stats}
    \centering
    \begin{tabular}{c|c||cc||cc|cc||cc|cc||cc|cc}
    \toprule
         \multicolumn{1}{c}{}& & \multicolumn{2}{c||}{\textbf{Original}} & \multicolumn{2}{c}{\textbf{Non-OOV}} & \multicolumn{2}{c||}{\textbf{OOV}} &
                \multicolumn{2}{c}{\textbf{Non-CG}} & \multicolumn{2}{c||}{\textbf{CG}} &
                \multicolumn{2}{c}{\textbf{DA Non-CG}} & \multicolumn{2}{c}{\textbf{DA CG}} \\
        \midrule
        \multirow{3}{*}{ Subset size } & 
        Train & \multicolumn{2}{c||}{50627}
                &\multicolumn{2}{c|}{36287} & \multicolumn{2}{c||}{36286}
                &\multicolumn{2}{c|}{29750} &\multicolumn{2}{c||}{32651}
                &\multicolumn{2}{c|}{36064 } &\multicolumn{2}{c}{36064} \\
        & Dev & \multicolumn{2}{c||}{8690}
                & \multicolumn{2}{c|}{6216}  & \multicolumn{2}{c||}{6216} 
                &\multicolumn{2}{c|}{3302} &\multicolumn{2}{c||}{3610}
                &\multicolumn{2}{c|}{6192} &\multicolumn{2}{c}{6192} \\
        & Test & \multicolumn{2}{c||}{13078}
                & \multicolumn{2}{c|}{3444}   & \multicolumn{2}{c||}{3444}  
                &\multicolumn{2}{c|}{1480} &\multicolumn{2}{c||}{1480}
                &\multicolumn{2}{c|}{3500} &\multicolumn{2}{c}{3500} \\
        \midrule
        \multirow{2}{*}{Action overlap}  
            % & Train-Dev     & 45 & 0.00 & 45 & 0.00 & 26 & 0.00     &  43   & 0.00      & 40    & 0.00 & 62 & ? & 43 & ? \\
            & Train-Test    & 46 & (1.00) & 22 & (0.71) & 2  & (0.32)     &  44   & (0.99)      & 39    & (0.83) & 39 & (0.97) & 35  & (0.93) \\
            & Dev-Test      & 45 & (0.99) & 21 & (0.71) & 2  & (0.33)     &  43   & (0.98)      & 38    & (0.83) & 38 & (0.96) & 37  & (0.92) \\
        \midrule
        \multirow{2}{*}{Scenario overlap}
            % & Train-Dev     & 18 & 0.00 & 18 &  0.00 & 18 & 0.00    &  18   & 0.00      & 16    & 0.00 & 12 & ? & 12 & ? \\
            & Train-Test    & 18 & (1.00) & 13 &  (0.83) & 13 & (0.83)    &  18   & (1.00)      & 16    & (0.84) & 12 & (1.00) & 12 & (1.00) \\
            & Dev-Test      & 18 & (1.00) & 13 &  (0.81) & 13 & (0.81)    &  18   & (1.00)      & 16    & (0.85) & 12 & (1.00) & 12 & (1.00) \\
        \midrule
        \multirow{2}{*}{Intent overlap}   
            % & Train-Dev     & 59 & 0.00 & 59 & 0.00  & 48 &  0.00   & 56    & 0.00      & 42    & 0.01 & 64 & ? & 44 & ? \\
            & Train-Test    & 59 & (1.00) & 23 & (0.88)  & 0  &  (0.00)   & 57    & (0.99)      & 41    & (0.39) & 27 & (0.74) & 0 & (0.00) \\
            & Dev-Test      & 58 & (0.99) & 22 & (0.87)  & 0  &  (0.00)   & 55    & (0.97)      & 39    & (0.38) & 20 & (0.56) & 0 & (0.00) \\
        \bottomrule
    \end{tabular}
\end{table*}
%%% Data stats %%%

\section{SLURPFOOD}
SLURP consists of 72395 recordings of 16520 utterances. Each sentence has 5 types of annotations: transcript, named entities, part-of-speech tags, scenario, and action. \emph{Scenario} is a label that indicates the general situation or context for the sentence, for example `qa' (question answering), `audio', or `transport'. The \emph{action} indicates what the speaker wants the (supposed) system to do (although often not very precisely), for example `factoid' in the `qa' scenario, `volume\_up' in the `audio' scenario, or `taxi' in the `transport' scenario. There is also a sixth, redundant annotation called \emph{intent}, which is the combination of scenario and action. For more details about SLURP, see \cite{bastianelli-etal-2020-slurp}.

We propose four distinct split configurations to assess generalisation in SLU. The mismatched acoustic environment splits are applicable to the scenario and intent classification tasks. The CG splits can be applied to scenario classification or be adapted for intent classification by predicting the scenarios and actions separately, for example by using a multi-task model. The OOV and Double-Action CG (DA CG) splits can be used only for the scenario classification task because their test sets include some action classes that are not in the training sets. The statistics of the splits are presented in Table \ref{tab:split_stats}.

\textbf{OOV splits.}
To create the OOV splits, we first select a test set, shared between the OOV and non-OOV portions. In this task, the test subset is sampled from the original SLURP test set, ensuring there are enough different intents.
To create the OOV training and dev subsets, we removed the intents present in the test set from the original training and dev subsets. This allows us to assess the model's OOV generalisation ability when the scenario is seen in training, but the task (action) is novel. For instance, the training set may contain `launch super mario' where the scenario is `play' and the action is `game'. Conversely, the test set could include an instruction `please play the next episode in the podcast', maintaining the `play' scenario, but in a combination with a novel `podcast' action.

To create the non-OOV training and development subsets to contrast with the OOV subsets, we replaced half of the difficult splits with elements containing intents present in the test set. While adding the new samples, we made sure that the scenario distribution remained the same as in the OOV split.

\textbf{CG splits.} \label{sec:dbca}
To generate the CG splits, we utilise the \emph{distribution-based compositionality assessment} (DBCA) \cite{keysers2019measuring} framework. DBCA is a method to determine how compositional two datasets are by calculating the divergences of the distributions of \emph{atoms}, primitive elements, on one hand, and the divergences of the distributions of combinations of atoms, called \emph{compounds}, on the other hand. To create a train-test split that requires compositional generalisation, the atom divergence should be similar between train and test splits, while compound distribution should diverge. In our CG splits, atoms are defined as the scenario and action labels, and the compound in each utterance is the combination of the two, i.e. the intent.

Similarity between distributions $P$ and $Q$ is calculated in the same way as in \cite{keysers2019measuring}, using the Chernoff coefficient $C_\alpha(P \Vert Q) = \sum_{k} p_k^\alpha \, q_k^{1-\alpha} \in [0, 1]$ \cite{chung1989measures}, with $\alpha=0.5$ for the atom distribution similarity and $\alpha=0.1$ for the compound distribution similarity. The \emph{divergence} is defined as $1-C_\alpha$.
A greedy algorithm, similar to that by \cite{moisio-etal-2023-evaluating}, splits the data aiming to maximise the train-test compound divergence while minimising the atom divergence. A non-CG split with minimal compound divergence, as well as minimal atom divergence, is also created to contrast with the CG split. In Table~\ref{tab:split_stats}, we can see that for the CG splits, the distributions of the actions and scenarios are relatively similar between training and test sets, whereas the distributions of intents diverge.

\textbf{Double-Action CG splits.}
We provide a second split configuration for testing the CG. In this split, the task is to predict the scenario from a combination of two utterances with different actions but same scenario. Again, for both the CG and non-CG versions, we used a unified test set and modified the training and development portions accordingly. The train, development, and test portions are derived from the original SLURP splits.

For the CG split, we ensured that no combination of the same two actions present in the test set appears in the training or development splits. To create the non-CG split, we removed half of the samples from the difficult set and replaced them with two action combinations present in the test set, preserving the class distribution as in the CG split.

\textbf{Microphone mismatch splits.}
In the original SLURP dataset, the utterances were captured either through a headset or from a distance, simulating diverse acoustic environments. We used this information to create splits for assessing SLU models' ability to generalise to mismatched acoustic environments. More specifically, we selected only the headset recordings from the official SLURP train and development splits. To test the models' classification ability in different acoustic environments, we created two test splits, derived from the official SLURP test set. The first one contains recordings made with a headset and the second one contains recordings made without a headset. We also made sure that the speakers were identical in both versions. The number of samples in training, development, and test splits is 24209, 4173. and 5552, respectively. The rest of the statistics are identical to the original splits (see Table \ref{tab:split_stats}).

% Description of the baseline systems
\section{Experiments and results}
\subsection{Baseline systems}

The primary goal of the proposed splits is to have resources for testing the generalisation ability of E2E SLU models. To test the generalisation ability of the models on the proposed splits, we created baseline E2E SLU systems for each of the splits and trained them on the scenario classification task.

All four split configurations are designed for the sequence classification task. To do sequence classification, we fine-tuned the English pre-trained WavLM-base-plus model \cite{chen2022wavlm} (90.2M trainable parameters) by adding a classification layer and training it with the negative log-likelihood loss. We chose this model due to its superior performance on various speech-related tasks. To combine the utterances in the DA CG/non-CG splits, we processed them individually using the WavLM model. The embeddings of the two utterances are then averaged to get a single vector representation.

All the models were trained with a single V100 GPU for 30 epochs, using a batch size of 10. In the OOV, CG and Mic mismatch, training for one epoch with the WavLM model took around 16 minutes, whereas for the DA CG it took 45 minutes. For more details about the hyperparameters, refer to the provided code repository.

%%% Results table %%%
\begin{table}[h!]
    \caption{Micro F1 scores for the scenario classification task. The results are obtained by averaging the scores from two runs with different seeds. The 95\% confidence intervals calculated with the bootstrap method are given in the brackets.}
    \label{tab:baseline_results}
    \centering
    \begin{tabular}{|c|c|c|c|}
      \toprule
      \textbf{Split} & \textbf{WavLM} & \textbf{TOPK} & \textbf{Segment} \\
      \midrule
      Non-OOV & \textbf{85.6 (85, 87)} & 84.1 (83, 86) & 85.0 (84, 87) \\
      OOV & 52.7 (51, 55) & \textbf{54.1 (53, 56)} & 53.4 (52, 55) \\
      \cmidrule{1-4}
      Non-CG & 80.9 (79, 83) & 80.9 (79, 83) & \textbf{81.8 (80, 84)} \\
      CG & \textbf{61.1 (59, 64)} & 57.6 (56, 60) & 59.4 (57, 62) \\
      \cmidrule{1-4}
      DA Non-CG & 88.9 (88, 90) & 88.7 (88, 90) & \textbf{89.0 (88, 90)} \\
      DA CG & 82.6 (81, 84) & \textbf{87.4 (87, 89)} & 86.3 (85, 88) \\
      \cmidrule{1-4}
      Mic Headset & 85.4 (85, 86) & \textbf{85.6 (85, 87)} & \textbf{85.6 (85, 87)} \\
      Mic Other & \textbf{69.9 (69, 71)} & 63.7 (63, 65) & 69.2 (68, 71) \\
      \bottomrule
  \end{tabular}
\end{table}
%%% Results table %%%

% Results section
\subsection{Results}
\iffalse
\begin{table*}[t!]
    \caption{Top 3 most important words for the WavLM predictions, selected using the feature ablation and integrated gradients techniques.}
    \label{tab:important_words}
    \centering
    \begin{tabular}{|c|c|c|c|c|}
        \toprule
        & \multicolumn{2}{c|}{\textbf{Feature ablation}} & \multicolumn{2}{c|}{\textbf{Integrated gradients}} \\
        \cmidrule{2-5}
        \multirow{1}{*}{Split: Scenario-Action} & Easy splits & Difficult splits & Easy splits & Difficult splits \\
        \midrule
        \textbf{alarm\_query} & {set, are, alarms} & {set, alarms, are} & {alarms, alarm, set} & {alarms, my, alarm} \\
        \textbf{iot\_coffee} & {make, coffee, machine} & {coffee, make, me} & {coffee, make, a} & {coffee, make, me} \\
        \textbf{music\_dislikeness} & {again, play, this} & {song, again, play} & {hate, this, play} & {this, don't, hate} \\
        \textbf{play\_audiobook} & {book, my, to} & {the, my, audiobook} & {the, play, resume} & {play, the, of} \\
        \textbf{play\_game} & {play, a, me} & {a, me, chess} & {play, a, me} & {of, game, a} \\
        \textbf{recommendation\_movies} & {are, movies, with} & {good, movies, movie} & {movies, are, what} & {are, movies, what} \\
        \textbf{takeaway\_order} & {order, for, a} & {order, takeout, some} & {order, call, for} & {order, for, a} \\
        \textbf{transport\_taxi} & {to, call, book} & {me, call, book} & {me, to, taxi} & {to, a, me} \\
        \bottomrule
  \end{tabular}
\end{table*}
\fi

As an evaluation metric, we used the micro F1 score. Additionally, we calculated the 95\% confidence intervals using the bootstrap method\footnote{Ferrer, L. and Riera, P. Confidence Intervals for evaluation in machine learning [Computer software]. \url{https://github.com/luferrer/ConfidenceIntervals}}. The scenario classification results are presented in Table \ref{tab:baseline_results}.

On the OOV split, the baseline WavLM model performs significantly worse than the non-OOV, with a 32.9 percentage point performance degradation. The big performance gap between the easy and difficult splits suggests that the model struggles to generalise when presented with context unseen during training.

On the CG/non-CG splits the difference in performance is smaller but still noticeable. On these splits, the WavLM got a performance decrease of 19.8 points. When using a bigger context, like in the case of the DA CG/non-CG splits, the gap in performance is more narrow, with a decrease of 6.3.

When evaluating the baseline WavLM model on utterances with the same and different acoustic environments as those in training we can see again a large drop in performance. Namely, when presented with utterances that do not use a headset, the performance drops 15.5 percentage points. Since these splits do not require any modification to the architecture to perform intent classification, we trained the baseline on that task. The results showed that the performance dropped from 79.7\% to 42.1\% when presented with samples that do not use the headset.

% Prediction analysis
\subsection{Why do the models fail to generalise?}

\begin{figure}[!h]
  \centering
  \includegraphics[width=0.92\linewidth]{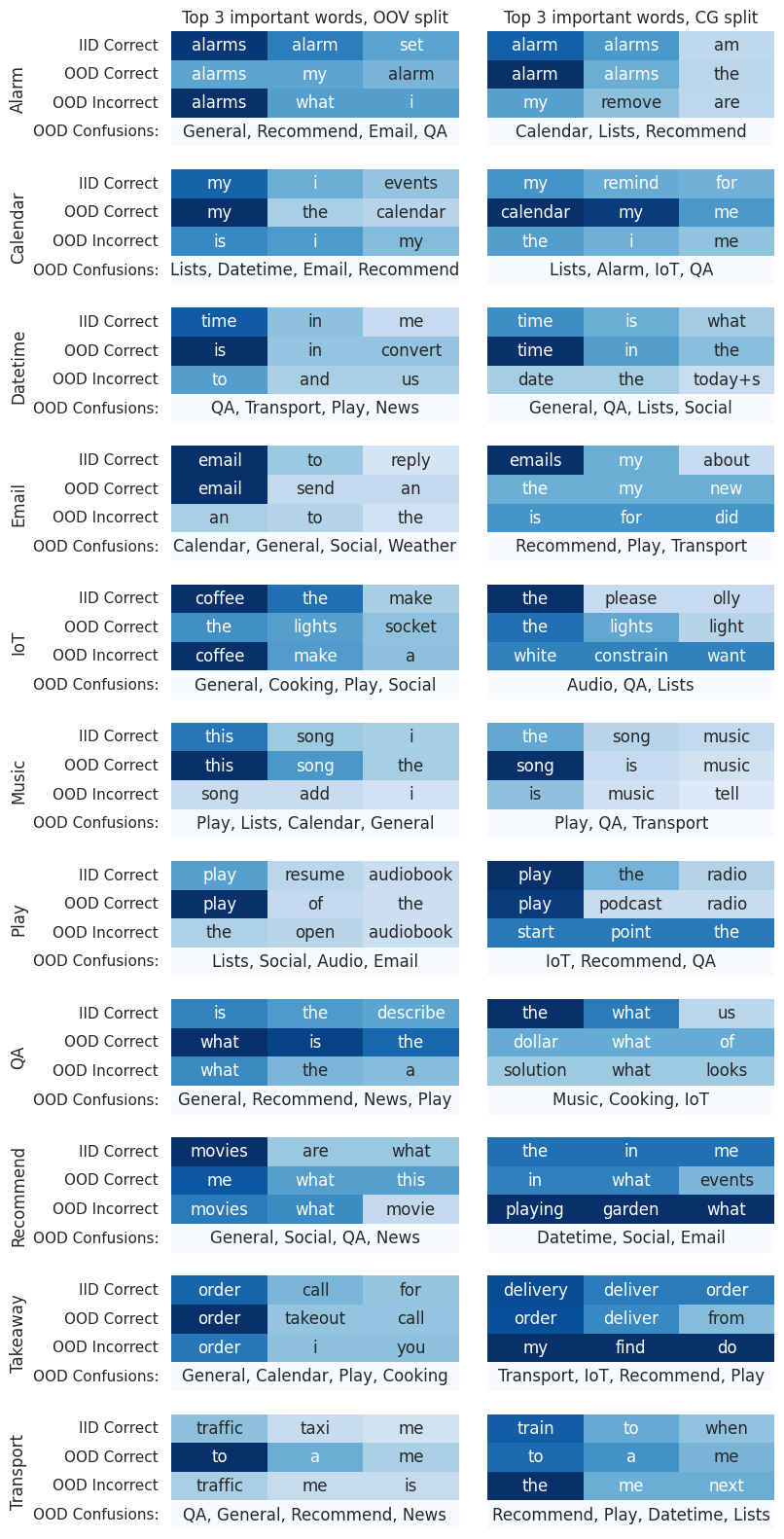}
    \caption{Top 3 words that are most frequently the most important word for the prediction (colour intensity signifies frequency), determined by the IG method, for the OOV and CG splits. The first two rows in each matrix show the most important words for the samples that were correctly predicted by the IID and OOD splits, respectively. The third row shows the most important words for the incorrectly classified samples. The last row shows the most commonly confused classes on the OOD split with respect to the true class.}
  \label{fig:topwords}
\end{figure}

As evident from the results, the WavLM model exhibits performance loss when evaluated on the OOD splits. This trend shows that even though the model has seen large amounts of data during pre-training, it still struggles to generalise. To investigate the reason why the model fails to generalise, we employed an interpretability technique called Integrated Gradients (IG) to determine the most important words for each prediction. IG is a technique for estimating the contribution of the input segments (in our case the word regions) to the final prediction. The attributions are calculated by creating interpolation steps from the baseline, in our case zeros, to the input and integrating along the steps. Then, integrated gradients are obtained by accumulating the gradients. For more technical details, refer to \cite{sundararajan2017axiomatic}. To obtain the most important word for each prediction, we took the word region with the highest attribution value.

Since we are interested in the contribution of each word to the final prediction, we need to align the audio signal with the corresponding transcript. For this, we used an ASR-fine-tuned wav2vec2 model. We collected the most important word for each classified sample and counted the frequencies of these words within each scenario class. Figure~\ref{fig:topwords} shows the 3 words most frequently determined as the most important by the IG method. We see the models often assign a high attribution value to stopwords, such as `a' or `the'. Focusing on stopwords may indicate that the model has learned spurious correlations between input features and output classes. For example, the models might learn to classify the samples based on syntactic structures independent of their semantic content, which would lead to an inability to generalise to novel utterances that are structured differently but have the same intent, or vice versa.

Comparing the important words for correct predictions in the CG split to those in the the IID (non-OOD) split (right-hand-side column of matrices, the 1st and 2nd row in each matrix), there is a tendency that the most important words for the CG split are semantically more relevant to the class than those for the IID split. For example, the word `calendar' in the calendar class, or `song' in music class, `lights' in the IoT class, or `podcast' in the play class. (Although there are 2 counterexamples: email and transport classes.) This could mean that in the CG split, the model has to use the semantically relevant words to predict the class, whereas in the IID split, the model can more often use stopwords to give correct classification. We take this to mean the CG split more readily assesses the models' capacity to learn to use features that help generalising to novel samples.

Comparing the important words for correct predictions to those for incorrect predictions can also give some clues as to why the model confuses some classes. For example, the Music class seems to be confused with the Play class because of the word `song' in the OOV split, and `music' in the CG split.

% Improving the generalisation
\subsection{Towards improving the generalisation}
Hoping to improve the generalisation of the WavLM model, we experimented with two approaches: TOPK \cite{kawaguchi2020ordered} and segmented processing. 
TOPK presents a straightforward method that preserves the existing architecture without any modifications. It involves averaging solely the top-k highest losses within each mini-batch, rather than all individual losses. Through experimentation on a subset of the data, we explored values of k ranging from 2 to 6 to select the optimal parameter. The best results were obtained with a k=2, therefore we use that value in the results reported in Table \ref{tab:baseline_results}.

For the second approach, we apply segment-based processing by obtaining the embedding vector from WavLM and splitting it along the temporal dimension in segments with overlapping windows. Then, we take the mean of each segment to get a single embedding per segment. Finally, we compute the mean of the embedding segments and pass it through the output layer. If the utterance is too short and can not be segmented, we use the whole, without segmenting. To determine the most optimal segment size, we conducted experiments by training on a subset of the training data. We experimented with segment sizes of 20, 40, 60, 80, 100, 120, and 140. The stride is always half of the segment size. We obtained the best results with a segment size of 120 and a stride of 60.

The result presented in Table~\ref{tab:baseline_results} show that both approaches improve the generalisation on some of the splits, but not on all of them. For instance, for the OOV splits, both TOPK and segmented processing produce better results than the baseline, with TOPK being better. On the CG splits generated with the DBCA method, both TOPK and segmented processing failed to improve the generalisation. However, for the DA CG split, both approaches outperformed the baseline. On the Mic other split, the TOPK approach performs significantly worse than the segmented processing. One reason for that could be due to the training set containing only samples recorded with a headset that have very little variance of the acoustic conditions, so the samples with highest losses are probably difficult not due to the acoustic environment but due to other factors instead.

% Limitations
\section{Limitations}
% limitations of data splits
In this study, we mostly focus on the scenario classification task, although intent classification is another task made possible by the intent labels in the SLURP data. Similarly, named entity annotations in the SLURP dataset could also be used to create a different type of CG splits.
% limitations of interpretability
We aimed to shed light on the reasons for the poor generalisation results of the models by utilising the integrated gradients method, and we made some hypotheses based on the most important words, but these hypotheses are not yet conclusively tested, and therefore only tentatively suggest weaknesses of the systems.
% limitations of the models? not SOTA?

% Conclusions
\section{Conclusions}
We introduced data splits to assess different types of OOD generalisation of SLU models. These splits systematically test the OOV, CG, and acoustic environment aspects of OOD generalisation. The empirical findings showed significant challenges in generalisation for the WavLM model across the OOD splits, underscoring the necessity for alternative modelling strategies. We investigated TOPK and segmented processing techniques, revealing improvements in some types of OOD generalisation, although still failing to improve generalisation to other types of OOD data. By analysing the most important words affecting the predictions, we noticed that the models often give high importance to the stopwords, indicating that they might learn spurious correlations. Moreover, we found that the important words in general differ between the OOD split types, providing different challenges to the models, which is corroborated by the fact that the best results are achieved by different model types depending on the split. Specifically, the CG split seems to require the system to focus on semantically relevant words more often than other splits do. In the future, we plan to use these findings of the weaknesses of the systems to develop methods for improving the generalisation of E2E SLU models.

\section{Acknowledgements}
We are grateful for the Business Finland project LAREINA under Grant 7817/31/2022. The computational resources were provided by Aalto ScienceIT.

%\clearpage % to see when we run out of space for references
\bibliographystyle{IEEEtran}
\bibliography{mybib}

\begin{thebibliography}{10}
\providecommand{\url}[1]{#1}
\csname url@samestyle\endcsname
\providecommand{\newblock}{\relax}
\providecommand{\bibinfo}[2]{#2}
\providecommand{\BIBentrySTDinterwordspacing}{\spaceskip=0pt\relax}
\providecommand{\BIBentryALTinterwordstretchfactor}{4}
\providecommand{\BIBentryALTinterwordspacing}{\spaceskip=\fontdimen2\font plus
\BIBentryALTinterwordstretchfactor\fontdimen3\font minus \fontdimen4\font\relax}
\providecommand{\BIBforeignlanguage}[2]{{%
\expandafter\ifx\csname l@#1\endcsname\relax
\typeout{** WARNING: IEEEtran.bst: No hyphenation pattern has been}%
\typeout{** loaded for the language `#1'. Using the pattern for}%
\typeout{** the default language instead.}%
\else
\language=\csname l@#1\endcsname
\fi
#2}}
\providecommand{\BIBdecl}{\relax}
\BIBdecl

\bibitem{liu2021towards}
J.~Liu, Z.~Shen, Y.~He, X.~Zhang, R.~Xu, H.~Yu, and P.~Cui, ``Towards out-of-distribution generalization: A survey,'' \emph{arXiv preprint arXiv:2108.13624}, 2021.

\bibitem{pmlr-v202-ray-chowdhury23b}
J.~Ray~Chowdhury and C.~Caragea, ``Monotonic location attention for length generalization,'' in \emph{Proceedings of the 40th International Conference on Machine Learning}.\hskip 1em plus 0.5em minus 0.4em\relax PMLR, 2023, pp. 28\,792--28\,808.

\bibitem{zhou2023algorithms}
H.~Zhou, A.~Bradley, E.~Littwin, N.~Razin, O.~Saremi, J.~Susskind, S.~Bengio, and P.~Nakkiran, ``What algorithms can transformers learn? a study in length generalization,'' in \emph{The 3rd Workshop on Mathematical Reasoning and AI at NeurIPS'23}, 2023.

\bibitem{lugosch19_interspeech}
L.~Lugosch, M.~Ravanelli, P.~Ignoto, V.~S. Tomar, and Y.~Bengio, ``{Speech Model Pre-Training for End-to-End Spoken Language Understanding},'' in \emph{Proc. Interspeech 2019}, 2019, pp. 814--818.

\bibitem{palogiannidi2020end}
E.~Palogiannidi, I.~Gkinis, G.~Mastrapas, P.~Mizera, and T.~Stafylakis, ``End-to-end architectures for asr-free spoken language understanding,'' in \emph{ICASSP 2020-2020 IEEE International Conference on Acoustics, Speech and Signal Processing (ICASSP)}.\hskip 1em plus 0.5em minus 0.4em\relax IEEE, 2020, pp. 7974--7978.

\bibitem{fodor1988connectionism}
J.~A. Fodor and Z.~W. Pylyshyn, ``Connectionism and cognitive architecture: A critical analysis,'' \emph{Cognition}, vol.~28, no. 1-2, pp. 3--71, 1988.

\bibitem{hupkes2020compositionality}
D.~Hupkes, V.~Dankers, M.~Mul, and E.~Bruni, ``{Compositionality decomposed: How do neural networks generalise?}'' \emph{Journal of Artificial Intelligence Research}, vol.~67, pp. 757--795, 2020.

\bibitem{broscheit2022distributionally}
S.~Broscheit, Q.~Do, and J.~Gaspers, ``Distributionally robust finetuning bert for covariate drift in spoken language understanding,'' in \emph{Proceedings of the 60th Annual Meeting of the Association for Computational Linguistics (Volume 1: Long Papers)}, 2022, pp. 1970--1985.

\bibitem{ray23_interspeech}
A.~Ray, Y.~Shen, and H.~Jin, ``{Compositional Generalization in Spoken Language Understanding},'' in \emph{Proc. INTERSPEECH 2023}, 2023, pp. 750--754.

\bibitem{lake2018generalization}
B.~Lake and M.~Baroni, ``Generalization without systematicity: On the compositional skills of sequence-to-sequence recurrent networks,'' in \emph{International Conference on Machine Learning}.\hskip 1em plus 0.5em minus 0.4em\relax PMLR, 2018, pp. 2873--2882.

\bibitem{keysers2019measuring}
\BIBentryALTinterwordspacing
D.~Keysers, N.~Sch{\"a}rli, N.~Scales, H.~Buisman, D.~Furrer, S.~Kashubin, N.~Momchev, D.~Sinopalnikov, L.~Stafiniak, T.~Tihon, D.~Tsarkov, X.~Wang, M.~van Zee, and O.~Bousquet, ``Measuring compositional generalization: A comprehensive method on realistic data,'' in \emph{International Conference on Learning Representations}, 2020. [Online]. Available: \url{https://openreview.net/pdf?id=SygcCnNKwr}
\BIBentrySTDinterwordspacing

\bibitem{yao2022structural}
Y.~Yao and A.~Koller, ``Structural generalization is hard for sequence-to-sequence models,'' in \emph{Proceedings of the 2022 Conference on Empirical Methods in Natural Language Processing}, 2022, pp. 5048--5062.

\bibitem{lepori2023break}
M.~Lepori, T.~Serre, and E.~Pavlick, ``Break it down: Evidence for structural compositionality in neural networks,'' \emph{Advances in Neural Information Processing Systems}, vol.~36, 2023.

\bibitem{viglino19_interspeech}
T.~Viglino, P.~Motlicek, and M.~Cernak, ``{End-to-End Accented Speech Recognition},'' in \emph{Proc. Interspeech 2019}, 2019, pp. 2140--2144.

\bibitem{potamianos2003robust}
A.~Potamianos and S.~Narayanan, ``Robust recognition of children's speech,'' \emph{IEEE Transactions on speech and audio processing}, vol.~11, no.~6, pp. 603--616, 2003.

\bibitem{haeb2020far}
R.~Haeb-Umbach, J.~Heymann, L.~Drude, S.~Watanabe, M.~Delcroix, and T.~Nakatani, ``Far-field automatic speech recognition,'' \emph{Proceedings of the IEEE}, vol. 109, no.~2, pp. 124--148, 2020.

\bibitem{gupta2005t}
N.~Gupta, G.~Tur, D.~Hakkani-Tur, S.~Bangalore, G.~Riccardi, and M.~Gilbert, ``The at\&t spoken language understanding system,'' \emph{IEEE Transactions on Audio, Speech, and Language Processing}, vol.~14, no.~1, pp. 213--222, 2005.

\bibitem{gaspers2022temporal}
J.~Gaspers, A.~Kumar, G.~Ver~Steeg, and A.~Galstyan, ``Temporal generalization for spoken language understanding,'' in \emph{Proceedings of the 2022 Conference of the North American Chapter of the Association for Computational Linguistics: Human Language Technologies: Industry Track}, 2022, pp. 37--44.

\bibitem{serdyuk2018towards}
D.~Serdyuk, Y.~Wang, C.~Fuegen, A.~Kumar, B.~Liu, and Y.~Bengio, ``Towards end-to-end spoken language understanding,'' in \emph{2018 IEEE International Conference on Acoustics, Speech and Signal Processing (ICASSP)}.\hskip 1em plus 0.5em minus 0.4em\relax IEEE, 2018, pp. 5754--5758.

\bibitem{haghani2018audio}
P.~Haghani, A.~Narayanan, M.~Bacchiani, G.~Chuang, N.~Gaur, P.~Moreno, R.~Prabhavalkar, Z.~Qu, and A.~Waters, ``From audio to semantics: Approaches to end-to-end spoken language understanding,'' in \emph{2018 IEEE Spoken Language Technology Workshop (SLT)}.\hskip 1em plus 0.5em minus 0.4em\relax IEEE, 2018, pp. 720--726.

\bibitem{arora21_interspeech}
S.~Arora, A.~Ostapenko, V.~Viswanathan, S.~Dalmia, F.~Metze, S.~Watanabe, and A.~W. Black, ``{Rethinking End-to-End Evaluation of Decomposable Tasks: A Case Study on Spoken Language Understanding},'' in \emph{Proc. Interspeech 2021}, 2021, pp. 1264--1268.

\bibitem{bastianelli-etal-2020-slurp}
\BIBentryALTinterwordspacing
E.~Bastianelli, A.~Vanzo, P.~Swietojanski, and V.~Rieser, ``{SLURP}: A spoken language understanding resource package,'' in \emph{Proceedings of the 2020 Conference on Empirical Methods in Natural Language Processing (EMNLP)}, B.~Webber, T.~Cohn, Y.~He, and Y.~Liu, Eds.\hskip 1em plus 0.5em minus 0.4em\relax Online: Association for Computational Linguistics, Nov. 2020, pp. 7252--7262. [Online]. Available: \url{https://aclanthology.org/2020.emnlp-main.588}
\BIBentrySTDinterwordspacing

\bibitem{sundararajan2017axiomatic}
M.~Sundararajan, A.~Taly, and Q.~Yan, ``Axiomatic attribution for deep networks,'' in \emph{International conference on machine learning}.\hskip 1em plus 0.5em minus 0.4em\relax PMLR, 2017, pp. 3319--3328.

\bibitem{kawaguchi2020ordered}
K.~Kawaguchi and H.~Lu, ``Ordered {SGD}: A new stochastic optimization framework for empirical risk minimization,'' in \emph{International Conference on Artificial Intelligence and Statistics}.\hskip 1em plus 0.5em minus 0.4em\relax PMLR, 2020, pp. 669--679.

\bibitem{chung1989measures}
\BIBentryALTinterwordspacing
J.~Chung, P.~Kannappan, C.~Ng, and P.~Sahoo, ``Measures of distance between probability distributions,'' \emph{Journal of mathematical analysis and applications}, vol. 138, no.~1, pp. 280--292, 1989. [Online]. Available: \url{https://www.sciencedirect.com/science/article/pii/0022247X89903351}
\BIBentrySTDinterwordspacing

\bibitem{moisio-etal-2023-evaluating}
A.~Moisio, M.~Creutz, and M.~Kurimo, ``Evaluating morphological generalisation in machine translation by distribution-based compositionality assessment,'' in \emph{Proceedings of the 24th Nordic Conference on Computational Linguistics (NoDaLiDa)}.\hskip 1em plus 0.5em minus 0.4em\relax T{\'o}rshavn, Faroe Islands: University of Tartu Library, May 2023, pp. 738--751.

\bibitem{chen2022wavlm}
S.~Chen, C.~Wang, Z.~Chen, Y.~Wu, S.~Liu, Z.~Chen, J.~Li, N.~Kanda, T.~Yoshioka, X.~Xiao \emph{et~al.}, ``Wavlm: Large-scale self-supervised pre-training for full stack speech processing,'' \emph{IEEE Journal of Selected Topics in Signal Processing}, vol.~16, no.~6, pp. 1505--1518, 2022.

\end{thebibliography}
\end{document}